\documentclass[conference]{IEEEtran}
\usepackage{graphicx}
\usepackage{url}
\graphicspath{ {./images/} }

\usepackage{listings,multicol}  
\usepackage{filecontents} 

\ifCLASSINFOpdf
\else
\fi
\usepackage{caption}
\begin{document}
%
\title{RelExt: Relation Extraction using Deep Learning approaches for Cybersecurity Knowledge Graph Improvement}

\author{
\begin{tabular}{c} Aditya Pingle, Aritran Piplai, Sudip Mittal, Anupam Joshi  \\ University of Maryland, Baltimore County, Baltimore, MD 21250, USA \\ $\lbrace$adiping1, apiplai1, smittal1, joshi$\rbrace$@umbc.edu \end{tabular} \and
\begin{tabular}{c} James Holt, and Richard Zak \\ Laboratory for Physical Sciences \\ $\lbrace$holt, rzak$\rbrace$@lps.umd.edu \end{tabular} 
}

\maketitle

\begin{abstract}
Security Analysts that work in a `Security Operations Center' (SoC) play a major role in ensuring the security of the organization. The amount of background knowledge they have about the evolving and new attacks makes a significant difference in their ability to detect attacks. Open source threat intelligence sources, like text descriptions about cyber-attacks, can be stored in a structured fashion in a cybersecurity knowledge graph. A cybersecurity knowledge graph can be paramount in aiding a security analyst to detect cyber threats because it stores a vast range of cyber threat information in the form of semantic triples which can be queried. A semantic triple contains two cybersecurity entities with a relationship between them. In this work, we propose a system to create semantic triples over cybersecurity text, using deep learning approaches to extract possible relationships. We use the set of semantic triples generated through our system to assert in a cybersecurity knowledge graph. Security Analysts can retrieve this data from the knowledge graph, and use this information to form a decision about a cyber-attack.


\end{abstract}

\IEEEpeerreviewmaketitle

\section{Introduction}

Cyber-attacks aim to compromise the confidentiality, integrity, and availability of information. These attacks target individuals, small and medium enterprises. Today attacks such as a denial of services, malicious codes executed via a backdoor or trapdoors, malware, trojans are some of the most common attack types \cite{cyber_trends_oxford}. 
Every year researchers and cyber defense professionals find millions of attack and malware  variants in the wild \cite{symantec_report}.

Now a `Security Analysts' working in a `Security Operations Center' (SoC) needs to keep up with variants of cyber threat intelligence that's available online so as to secure their organization. These analysts need to process this information keeping in mind their local defensive setup.

Knowledge about cyber-attacks and malwares specifically their means, target, etc. is available in the `wild' as cyber threat intelligence. It is possible to develop a cyber informatics pipeline which scours the web for threat intelligence, extracts and mines knowledge from these intelligence samples and represents them as a scheme fit for Security information and event management (SIEM) systems. 

Threat intelligence sources are generally of two types, \textit{covert} and \textit{overt}. Overt or `open source intelligence' (OSINT) is available through various sources like, cybersecurity blogs, cybersecurity reports, CVE \cite{cvelist}, CWE \cite{CWE}, National Vulnerability Datasets \cite{nvd}, and any cybersecurity text publicly available.
Various SIEM systems, fetch, process, and store much of the cyber threat intelligence available through OSINT sources. 
Knowledge Graphs (KG) specifically to store cyber threat intelligence has been used by many cybersecurity informatics systems. These cybersecurity knowledge graphs can not only store but are also be able to retrieve data and answer complex queries asked by the security analysts.


The dependence of various cybersecurity informatics systems on various knowledge representation schemes makes it imperative that we develop systems that improve these representations. Some examples are the Intrusion Detection System Knowledge Graphs proposed by Undercoffer et al. \cite{undercofferIDS} and Unified Cybersecurity Ontology (UCO) \cite{unified_cyber_ontology}. In case of a cybersecurity knowledge graph, this boils down to improving the quality of semantic triples generated from various cyber threat intelligence sources. An ontology can be used to provide a system with base cybersecurity classes and the relations that exist between them.

Semantic triple generation is a key component in the Knowledge Graph population(See Section \ref{related work}). A semantic triple for cybersecurity comprises of a pair of cybersecurity entities and a relationship between the entities. A cybersecurity entity can be a word or a group of words extracted from cybersecurity text. Generally, a Named Entity Recognizer (NER) is used to extract entities that form the base knowledge graph. Multiple such cybersecurity domain specific named entity recognizer currently exist \cite{DBLP:journals/corr/BridgesJIG13}\cite{Jones:2015:TRE:2746266.2746277}\cite{Information_Extraction_of_Security_related_entities_and_concepts_from_unstructured_text_}\cite{Mulwad:2011:EIS:2052137.2052279}\cite{JIA201853}\cite{strauss-etal-2016-results}\cite{r-etal-2018-teamdl}.

The next step in cybersecurity knowledge graph creation is, predicting the right relationship between two specific entities extracted from a specific cyber threat intelligence source. 
As soon as we figure out the various relationships and entities, we can assert the semantic triple set in a knowledge graph. 

In this paper, we propose the \textit{RelExt} system that strives to improve various cyber threat representation schemes, especially cybersecurity knowledge graphs (CKG) by predicting relations between cybersecurity entities identified by cybersecurity named entity recognizer. These representation systems have been used to build various `Analyst Augmentation Systems' that aid a security analyst. These augmentation systems help keep an analyst working in an SoC environment updated about various developments in the wild that can impact the security of her organization. We believe that improving the base cyber threat intelligence representation will help improve the overall quality and performance of these analyst augmentation systems. 




RelExt is a feed-forward neural network model that predicts relationships between cybersecurity entities, that form up a triple. Existing models use features such as length of entities involved, words between the entities as features for the model. We leverage the contextual similarity between the entities to find out if the entities would make up a triple. In RelExt, we use contextual vector representation of cybersecurity entities identified by a cybersecurity specific NER.

The rest of the paper has been organized as follows - Section \ref{related work} discusses some related work. Section \ref{sec ckg} talks about building an ontology for our Cybersecurity Knowledge Graph. We introduce the RelExt system in Section \ref{sec RelExt}. Section \ref{sec experimental setup and evaluations} talks about the evaluation metrics. We conclude the paper in section \ref{sec conclusion}.

\section{Related Work}
\label{related work}


\subsection{Knowledge Graphs for cybersecurity} \label{systems}
Data management and exploration were for long supported by either keyword-based search technology or various stochastic matching approaches \cite{raskin}. Content and knowledge-based approach in contrast to the former approach, add a layer of sophistication. 
Often termed as the specification of a conceptualization, Ontology forms the exclusive way to represent and communicate facts and relations with multiple agents \cite{gruber1993}. Ontology is a set of classes with attributes and relationships between instances of various classes.
One of the earliest available ontology for the cybersecurity, which was built as an Intrusion Detection System (IDS) was introduced by Undercoffer et al. \cite{undercofferIDS}.
Unified Cybersecurity Ontology, an extension of the IDS ontology, built by Syed et al. \cite{unified_cyber_ontology} is a more cybersecurity domain focused ontology based on STIX 1.2 \cite{stix1doc}. We in this paper, extend some of the concepts from UCO 1.0, like various classes and relationships. 
Yan Jia et al. \cite{JIA201853} discuss knowledge graph population approaches for the cybersecurity domain. The authors use various machine learning approaches to extract entities of interest from the cybersecurity domain. Relationships are then predicted between these entities by calculating formulas and path ranking algorithms.
CyberTwitter \cite{CyberTwitter_Using_Twitter_to_generate_alerts_for_Cybersecurity_Threats_and_Vulnerabilities} and Cyber-All-Intel \cite{Thinking_Fast_and_Slow_Combining_Vector_Spaces_and_Knowledge_Graphs} are other systems that create a cybersecurity knowledge graph. 

Knowledge graphs consist of semantic triples which have a subject, predicate, and an object. A predicate is a relationship between the subject and the object. The semantic triple generation would thus require a relationship extractor. Alternatively, an existing knowledge graph is improved by looking for more relationships, or links, between entities, nodes, already available in the knowledge graph. Our paper concentrates on developing \textit{RelExt}, which uses a feed forward neural network to predict relationships between subject-object entity pairs and subsequently improve the knowledge graph population. 

\subsection{Relationship Extraction}

Relationships connect two entities found to be interesting by the system. Relationships could be symmetrical or asymmetrical. Entities can also end up with no relation to each other, depending upon the underlying ontology. Relationship extractors always end up with a lack of sufficient data to be trained with. Defining the relation might seem to be a straightforward task, but applying a relationship for two entities is an ambiguous task. Often, a difference of opinions and perspective to what a relationship means, end up with higher inner-annotation disagreement \cite{pawarRelSurvey}. 
Based on the approach, relation extraction can be classified as global level or mention level relation extraction \cite{pawarRelSurvey}. Mention level extraction expects a pair of entities to find a relationship for. Whereas, global level extraction only expects the corpus and ends up listing entities with the relationship. 
Relationship extractors can be either binary or n-ary classifiers. Binary classifiers try to predict whether a specific relation R holds between two entities. Whereas n-ary classifiers try to predict whether two entities hold one of the relationships from the predefined set \cite{bachRelSurvey}.

Relation extraction can be classified down to three approaches, supervised, semi-supervised, and unsupervised. Supervised approaches are explored by Kambhatla et al. \cite{kambhatla} and Zhao et al. \cite{zhao}. Some of the well known semi-supervised approaches include the DIPRE system introducted by Brin \cite{brin_dipre}, Snowball \cite{snowball_2000}, and Knowitall \cite{Etzioni:2004:WIE:988672.988687}. Hasegawa \cite{hasegawa_clustering} proposed one of the earliest approach for unsupervised clustering for relationship extraction. Knowitall is extended well to incorporate an unsupervised approach for relationship extraction by Rosenfeld et al. \cite{feldman_nonclustering}.

\subsection{Knowledge Graph Improvement}
Relation prediction or extraction is one elegant way to fill in missing links between entities of interests in a knowledge graph. A simple link prediction would not suffice this task, as nodes in KG carry the certain identity and the edges mean a certain connection between the identities \cite{yankaiKGC}. Moreover, we need to know if a relationship exists between pairs of entities along with the type of relationship.
Relationships between two entities, or between pair of the group of entities, can be classified based on their complexity and number of entities mapped. 
TransE \cite{transe_bordes} model predicts one-to-one relationships \cite{transh} based on the vector space for head and tail entities. 
TransH model, an improvement over TransE \cite{transh}, based over translating vectors on hyperplane \cite{transh} works on predicting many-to-many relationships. 
The TranSparse model uses sparse matrices to address the heterogenity and imbalance of knowledge graph which was missed out by TransE, TransH, TransR, and TransD models \cite{adap_sparse_mat}.
Our model, RelExt, is trained over vectors, specifically in the domain of cybersecurity, to extract relationships between pairs of cybersecurity entities.

\section{Cybersecurity Knowledge Graph}
\label{sec ckg}
 A use-case of our system is to aid in the development of a cybersecurity knowledge graph (CKG). A CKG generally has two main parts, a schema and various cybersecurity-related semantic triples. In order to build a CKG, the first thing we need to do is to define a schema. Next, we use our system to extract relationships between cybersecurity entities, which are words or groups of words, extracted from cybersecurity text. Once we have the entity-relationship set, we can populate the data consistent with the schema. 

Syed et al. in their paper \cite{unified_cyber_ontology} have developed an ontology for cybersecurity called Unified Cybersecurity Ontology (UCO 1.0). Their paper describes a prototype version of an ontology which encompasses various cybersecurity elements. The paper provided us with a starting point for the development of our system, as it mentions some classes and probable relationships that could exist between various classes. 

Members of our research group have worked previously on creating cybersecurity knowledge graph from different open source intelligence (OSINT) sources \cite{CyberTwitter_Using_Twitter_to_generate_alerts_for_Cybersecurity_Threats_and_Vulnerabilities, Thinking_Fast_and_Slow_Combining_Vector_Spaces_and_Knowledge_Graphs}. In this paper, we create a system `RelExt' that improves such knowledge graphs by predicting relations between various cybersecurity entities, as they occur in the text. RelExt (See Section \ref{sec RelExt}) leverages a deep neural net to automatically extract various relationships between pairs of entities to improve our CKG. 

\begin{figure}
\begin{center}
\includegraphics[width=0.5\textwidth]{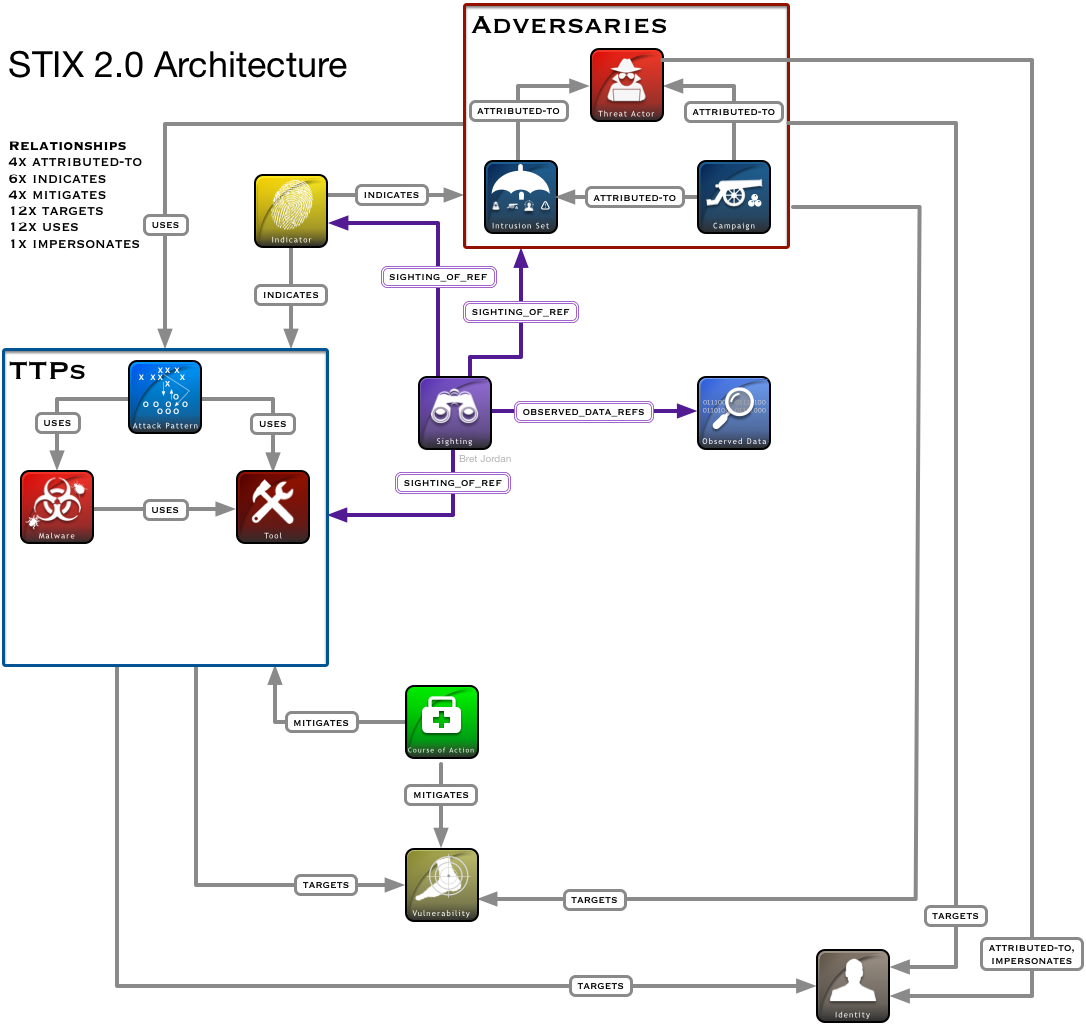}
\end{center}
\renewcommand{\baselinestretch}{1}
\small\normalsize
\caption{Relationships and some classes defined in Structured Threat Information eXpression (STIX) 2.0 architecture \cite{stix2doc}.}
\label{stix relationships}
\end{figure}

Defining entities and relationships in a cybersecurity ontology are paramount for building a CKG. We use the class definitions proposed by UCO \cite{unified_cyber_ontology} to define our entities of interest. The class definitions and the relationship definition present in UCO 1.0 were based on Structured Threat Information eXpression (STIX) 1.2. We use the relationships and classes defined by the Structured Threat Information eXpression (STIX) 2.0 to update UCO 1.0 and create UCO 2.0. \cite{stix2doc}. Figure \ref{stix relationships}, provides an overview of the relationships between various entity types as defined in STIX 2.0.

\subsection{Classes in UCO 2.0}
Here we explain some important classes that are present in UCO 2.0:
\begin{itemize}
    \item \textit{Software} : An entity that relates to a piece of code usually used as tool such as Office or Unix operating system.
    \item \textit{Malware} : An entity that refers to malicious code and/or software which is inserted into a system.
    \item \textit{Indicator} : An entity that contains a pattern which helps the administrator to indicate an ongoing attack or malicious activity.
    \item \textit{Vulnerability} : An entity that refers to a patch of bug or weakness that could be exploited by ill-intended users.
    \item \textit{Course-of-action} : An entity that refers an action or set of actions that either prevents or responds to an attack.
    \item \textit{Tool} : An entity that refers to a legitimate software that can be used by threat actors for malicious activities.
    \item \textit{Attack-pattern} : An entity that refers to steps that could result in an active attack on an individual or group of users.
    \item \textit{Campaign} : An entity that refers to grouping of activities that could lead to a malicious attack.
\end{itemize}

\subsection{Relationships in UCO 2.0}
In UCO 2.0 we have defined the following relationship, based on the STIX 2.0 definitions \cite{stix2doc}.
\begin{itemize}
    \item \textit{hasProduct} : Relationship where the subject and object entities, or just the object entity belong to software class.
    \item \textit{hasVulnerability} : Relationship where the subject entity belongs to a software class and object entity belongs to a vulnerability class.
    \item \textit{mitigates} : Relationship where the subject entity belongs to a course-of-action class and object entity belongs to malware or campaign class, wherein the subject entity aims to lessen the damages caused by object entity.
    \item \textit{uses} : Relationship where the subject entity belongs to a campaign or malware class and object entity belongs to a tool or software class, wherein subject entity aims to leverage object entity to carry on an attack.
    \item \textit{indicates} : Relationship where the subject entity belongs to a indicator class and object entity belongs to a malware or campaign class, wherein indicates. demonstrates presence or after effects of a object entity
    \item \textit{attributed-to} : Relationship where the subject entity belongs to campaign or intrusion-set class and object entity belongs to threat actor class wherein subject entity is attributed to object entity.
    \item \textit{related-to} : Relationship wherein subject entity and object entity are related to each other.
    \item \textit{located-at} : Relationship where entities are inter dependent or become a part of.
\end{itemize}

\subsection {Baseline Knowledge Graph and missing relationships}

In existing cybersecurity knowledge graph systems (See Section \ref{systems}) the resulting knowledge graph can be improved. A knowledge graph can have incorrect relationships between two cybersecurity entities, or may not even assert a relationship. In such a case, we can say that there are a few `missing' relationships. Our system RelExt improves such knowledge graphs by validating relationships and asserting values for missing relationships.  

\begin{figure}
\begin{center}

\includegraphics[width=0.5\textwidth]{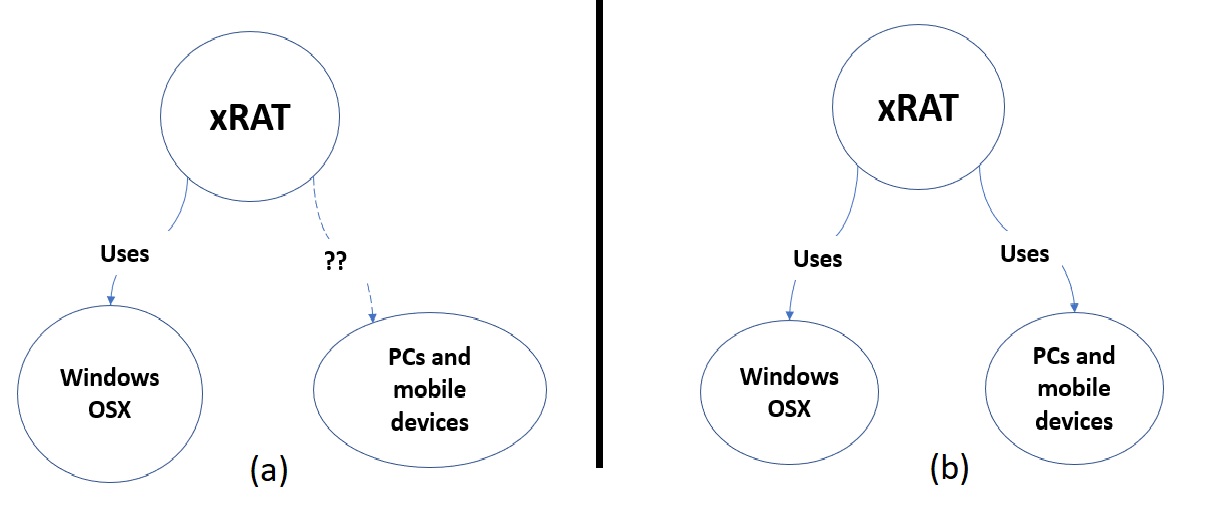}
\end{center}
\renewcommand{\baselinestretch}{1}
\small\normalsize
\caption{Missing Relationships: The diagram on the left depicts how a particular relationship is not asserted in the existing knowledge graph. The one on the right shows, the relationship asserted by RelExt.}
\label{RelExt_Missing}
\end{figure}

For example, in Figure \ref{RelExt_Missing}(a), there is a relationship which is not found by existing methods, hence we can say that this relationship is `missing' from the knowledge graph. Figure \ref{RelExt_Missing}(b), shows how RelExt can automatically suggest a value for the missing relationship. 

\section{RelExt: System Architecture}
\label{sec RelExt}

In this section, we introduce our RelExt system, which extracts relationships between two named entities. The input to RelExt is a pair of named entities extracted from a particular cybersecurity text. RelExt outputs an entity relationship set.

We use a Named Entity Recognizer (NER) to extract cybersecurity entities from a piece of cybersecurity text. This NER was created and used in the CyberTwitter system \cite{CyberTwitter_Using_Twitter_to_generate_alerts_for_Cybersecurity_Threats_and_Vulnerabilities}. The NER outputs a key-value pair, where the key is a cybersecurity entity and the value is a class as defined in UCO 2.0. A pair of such named entities serve as the input to RelExt. 

\begin{figure}
\centering
\includegraphics[width=0.5\textwidth]{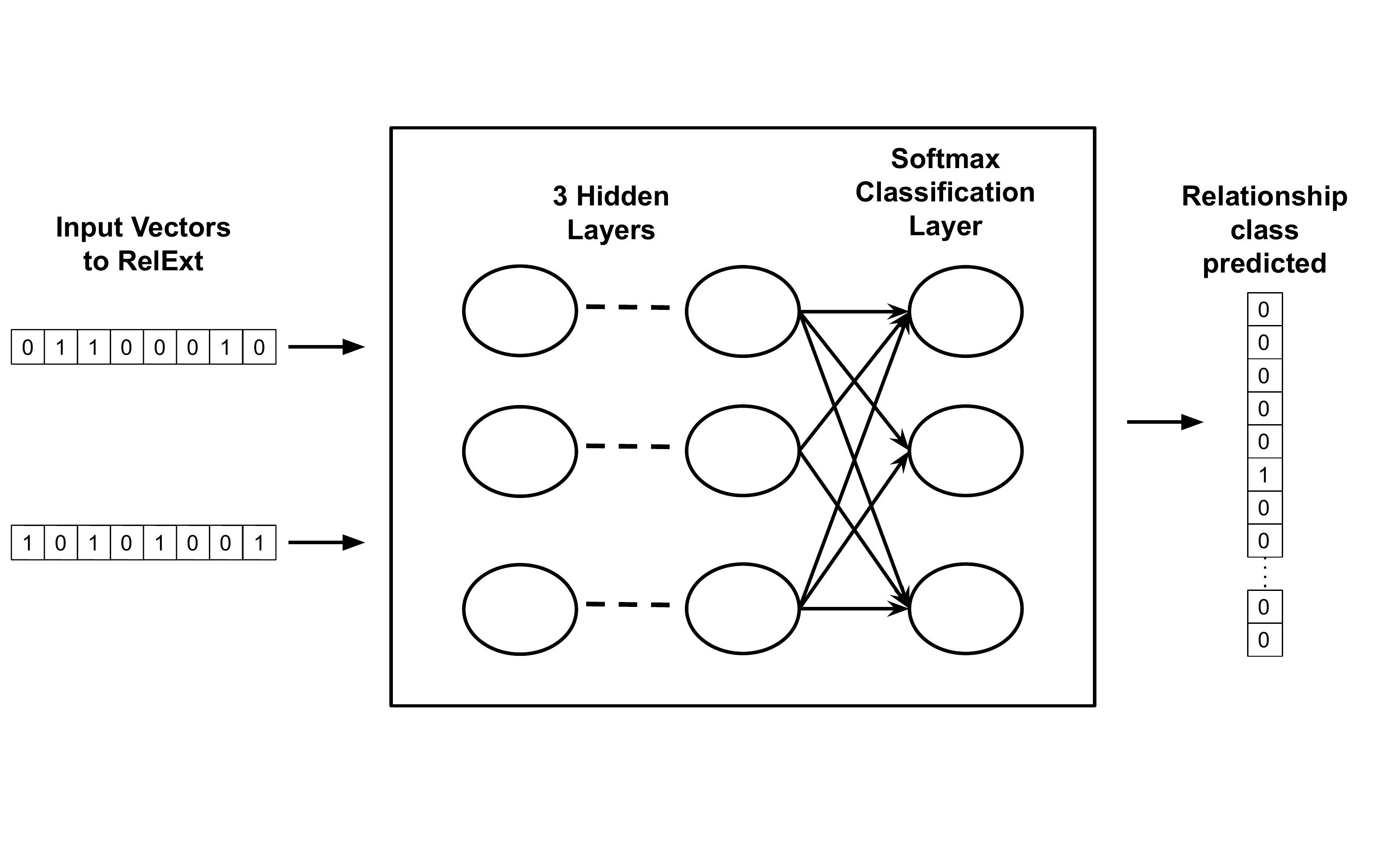}
\caption{Relationship Extractor - A Feed Forward Neural Network}
\label{RelExt}

\end{figure}

RelExt's input can be further processed based on our cybersecurity knowledge graph's schema. 
This allows us to remove some entity pairs which are not consistent with UCO 2.0 and STIX 2.0. For example, we do not provide the relationship extractor with an input where the entity pairs are of the type \textit{Campaign} and \textit{Version}, because our schema suggests there should be no relationship between them.
\begin{figure}
\includegraphics[width=0.5\textwidth]{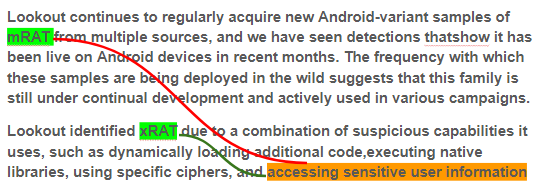}
\caption{The pair of entities  connected by the red line (\textit{mRAT}, \textit{accessing sensitive information}) is not provided as an input to RelExt as they are not in close proximity. The pair connected by green (\textit{xRAT}, \textit{accessing sensitive information}), however, is preserved.}
\label{Proximity Example}
\end{figure}
We also remove pairs of entities which are not in close proximity to one another in the pre-processing stage as explained in Figure \ref{Proximity Example}.
We have chosen a threshold of 35 words, which is used as a window-size limit. Two entities have to be within a window of 35 words, to be sent to the RelExt system for relationship prediction. We found that the average sentence length in the corpus was 14 words. We keep the window length as a function of the average length of sentences. We empirically chose 35 as the threshold for proximity. 
Having said that, this number is a heuristic, and it is open to experimentation. Users, of this system, can tune this parameter, depending upon their demand for precision and recall.

Once, we get the processed list of credible entity pairs, we can generate vector embeddings of these entities. We have another neural model, Word2Vec \cite{mikolov2013efficient}, which is trained on a cybersecurity corpus, which takes any particular word and then converts that word into a vector of fixed dimensions. We take two vectors, one for each entity, and provide them as an input to RelExt, which is a Feed-Forward Neural Network (FFNN) classifier. RelExt then predicts one out of the candidate relationships which we have assumed to exist between the named entity pairs from UCO. 

\begin{figure*}
\centering
  \includegraphics[scale=0.6]{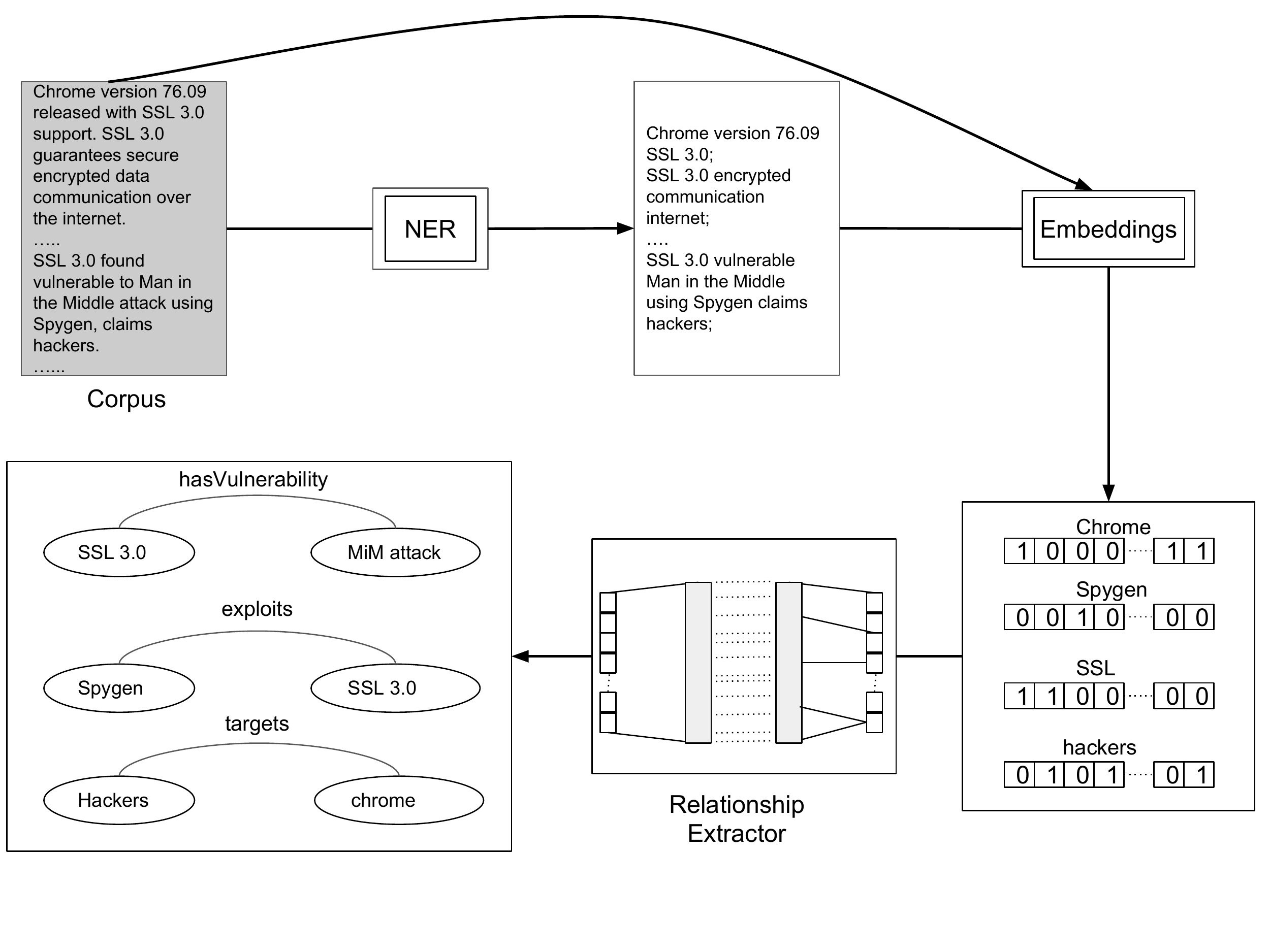}
  \caption{A cybersecurity document is processed by the NER. NER extracts the entities based on classes defined in UCO 2.0. Further, the vector representations of the entities are generated using a Word2Vec model. A preprocessed set of entity pairs act as input for the RelExt. Based on the relationships predicted for the entities by RelExt, we assert it in the Cybersecurity Knowledge Graph (CKG).}
     \label{system_architecture}
\end{figure*}

The model (See Figure \ref{RelExt}) has 3 hidden layers along with input and output layers. Hidden layers form the fully connected layers. Input layer accepts two entities in their vector representations. The neural model accepts the two resulting vectors of dimension 200. 


These vector embeddings are concatenated at the initial layer. The concatenated embeddings are propagated to the hidden layers. Hidden layers are non-linearly activated using the sigmoid function. Weights and biases are learned through training the neural model. The three hidden layers with 200, 100, 50 neurons respectively. The output is generated using a softmax layer.

\begin{figure*}
\centering
  \includegraphics[scale=0.6]{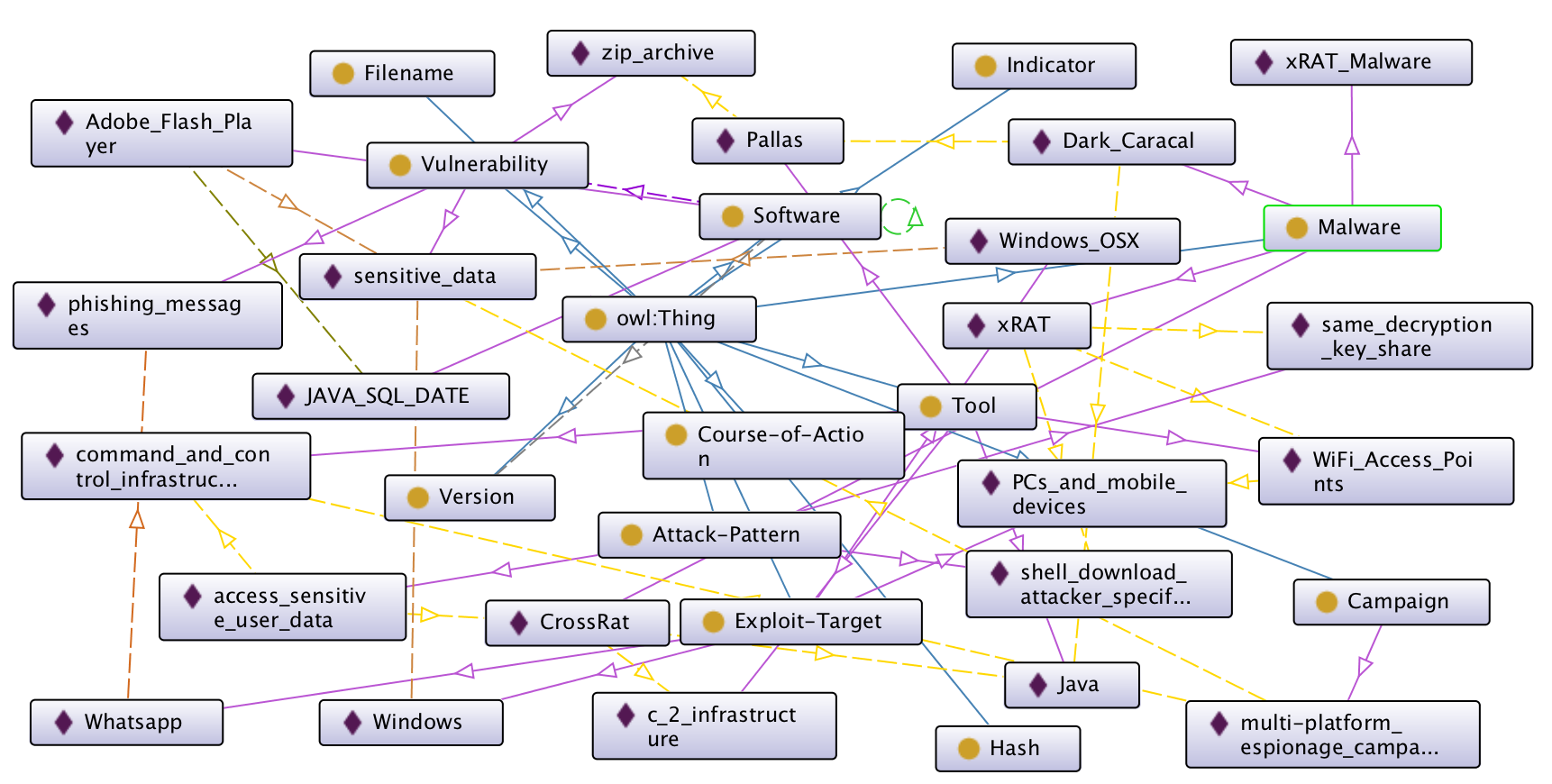}
  \caption{Asserted Knowledge Graph with relations added using our RelExt system.}
     \label{populated kg}
\end{figure*}

\subsubsection{Updating the CKG}\label{kgdc}
The output of RelExt is a relationship set for every pair of entities which are provided as input to the model. So, for every pair of extracted entities from the cybersecurity text, we predict a relationship, from the candidate relationships. 
We take this entity-relationship set and assert it in our cybersecurity knowledge graph. As an example, in Figure \ref{populated kg}, we showcase a part of our cybersecurity knowledge graph using a tool called Protege \cite{protege}. The asserted knowledge graph contains information captured from multiple sources. The example shows description of two malware attacks, \textit{Dark Caracal} and \textit{crossRAT}. We can see all the classes like Vulnerability, Course-of-Action, etc. and the individual entities of the classes which were found in the description. We also see the relationships asserted between the entity pairs, which are represented as lines connecting these entities. E.g., 'Wi-Fi Access Points' is a \textit{Tool} which is used by 'xRAT' \textit{Malware}. We can capture other information related to \textit{Softwares} mentioned in the cybersecurity text. For example, we see \textit{Software} `Adobe Flash Player' \textit{hasVulnerability} \textit{Vulnerability} `sensitive data'. We can also capture other information about the software, such as the other \textit{softwares} `Adobe Flash Player', in turn, uses. We can see \textit{Software} `Java SQL Date' is being used by `Adobe Flash Player'. We also see information about the \textit{exploit-targets} for respective \textit{malwares}. We see `Whatsapp' as an \textit{exploit-target} which \textit{hasVulnerability} `phishing messages'. `Windows OSX', `PCs and mobile devices' are other \textit{exploit-targets} asserted in the knowledge graph, which are affected by the `xRAT' Malware. Exploiting `PCs and mobile devices' can be seen as a part of a \textit{campaign} `multi-platform espionage campaign'. The information presented in the knowledge graph is useful in deriving similarities between two malware attacks. We can infer that both `Dark Caracal' and `xRAT' \textit{use} \textit{Tool} `Java'. Further information about which Tools `Dark Caracal' uses is also asserted in the knowledge graph. `Pallas' is another Tool, used by `Dark Caracal'. 

\section{Experimental Setup and Evaluations}
\label{sec experimental setup and evaluations}
In this section, we first describe our corpus and experimental setup. We then evaluate our RelExt system.

\subsection{Cybersecurity Corpus}
Our proposed system aims to find relations between cybersecurity entities. We construct our corpus from publicly available cybersecurity text. We have extracted information from cybersecurity bulletins, social media, National Vulnerability Datasets. Along with these publicly available information sources, we also collect detailed technical reports authored by cybersecurity specialists. 
Figure \ref{Proximity Example}, is an extract from one of these technical reports. 

However, sometimes this content is not available in raw text. The foremost action would be extracting raw text from these documents/reports. We have used open source libraries available on the internet to extract raw text from HTML, PDF, JSON, XML sources. 

We next describe our training set: 
\begin{itemize}

 \item Cybersecurity blogs, cybersecurity technical reports, are publicly available for use over the internet. We procured about 2 GigaBytes of data from publicly available. The data had about 474 detailed technical reports and blogs. The reports have detailed analysis of what security analysts and researchers have learned after studying an attack. Since they are very detailed and size-able reports, they provide us with a large number of relationships that exist between various cybersecurity entities. They are also structurally quite different from one another, as opposed to smaller pieces of cybersecurity texts like Common Vulnerabilities and Exposures (CVE), tweets, etc. This gives us an opportunity to build a generalized model. Having a diverse corpus will help the system to avoid bias. 


\item \textit{Common Vulnerabilities and Exposures (CVE) Corpus}: CVE JSON feeds have a textual description of the vulnerability present in a product. We collected approximately 90,000 JSON entities which are updated in the National Vulnerability Dataset \cite{cvelist}. 

\item \textit{Microsoft and Adobe Security Bulletins}: Microsoft and Adobe release security flaws and vulnerabilities in their respective software in the form of bulletins. Since these are publicly available and these two companies have produced applications which are ubiquitous, knowledge about them is valuable while detecting an attack. Thus, we also use them in our training set for relationship extraction.

\begin{table}
\begin{center}
 \begin{tabular}{||c c||} 
 \hline
 Splits & Accuracy \\ [0.5ex] 
 \hline\hline
 80-20 & 96.21\% \\ 
 \hline
 70-30 & 91.88\% \\
 \hline
 66-34 & 91.34\% \\
 \hline
\end{tabular}
\caption{Accuracy for various splits on training and testing data.}
\label{splits table}
\end{center}
\end{table}

\begin{table}
\begin{center}
 \begin{tabular}{||c c c c||} 
 \hline
 Relationship Classes & Precision & Recall & F-1 Score \\ [0.5ex]
 \hline
 hasProduct & 49 & 97 & 65 \\ 
 \hline
 hasVulnerability & 92 & 74 & 82 \\ 
 \hline
 uses & 100 & 88 & 93 \\ 
  \hline
 indicates & 80 & 90 & 85 \\ 
  \hline
 mitigates & 55 & 70 & 62 \\ 
  \hline
 related-to & 92 & 74 & 66 \\ [1ex] 
 \hline
\end{tabular}
\caption{Precision, Recall, and F-1 score for relationship classes.}
\label{precision table}
\end{center}
\end{table}

We can extract information such as \textit{Software} entities that would help us build the \textit{hasProduct} relationship. Moreover, we can also process the problems possessed by the product or a specific version of the product. The problem description contains information about the weaknesses or vulnerabilities known for the product. This information is sometimes represented by Common Weakness Enumeration (CWE) IDs \cite{CWE}. We have also processed the CWE-IDs made available by MITRE \cite{MITRE}. Whenever a CWE-ID is met, we try to get a description of the CWE-ID. Problem description ends up being the Vulnerability entities. These entities could form up a \textit{hasVulnerability} relationship with Software entities.

\item \textit{STIX Corpus}: 
We also train our relationship extraction model with the semantic triples generated using the information from Trusted Automated eXchange of Indicator Information (TAXII) \cite{TAXII} servers. These triples are extracted in the source, relation, target format and processed.

\end{itemize}

Next we use this corpus to generate our Feed Forward Neural Network based RelExt system and evaluate it.

\subsection{Evaluations}

We train RelExt with relationships from the aforementioned corpus. Along with manually annotated relationships by cybersecurity experts and relationships that were extracted from the corpus, we have a training set that contains 33,000 relationships. The vector representations are generated using a pre-trained Word2Vec model, over our cybersecurity corpus. The training set is iterated over in a batch size of 100 with 50 epochs. Each element of the batch is a pair of vectors for subject and object entity.

We perform training with various splits like: 80-20, 70-30, and 66-34 and the accuracy metrics for these splits have been presented in Table \ref{splits table}. Table \ref{precision table} shows the precision and recall values for various relationship classes.


\begin{table}
\begin{center}
 \begin{tabular}{||c c c||} 
 \hline
 Document  & Entity Pairs  & Relations Predicted \\ [0.1ex] 
 \hline
 Dark Caracal & 1287 &  1127 \\ 
 \hline
 CrossRat & 123 &  109 \\[1ex] 
 \hline
\end{tabular}
\caption{Number of Relationships predicted from the malware descriptions of Dark Caracal and CrossRAT malwares.}
\label{relationships in docs}
\end{center}
\end{table}

\begin{table}
\begin{center}
 \begin{tabular}{||c c||} 
 \hline
 Relationship Classes & Predicted \\ [0.5ex] 
 \hline
 hasProduct &  340\\ 
 \hline
 hasVulnerability & 93\\ 
 \hline
 uses & 307\\[1ex] 
 \hline
\end{tabular}
\caption{Various relationship classes found in Dark Caracal and CrossRat malware descriptions.}
\label{relationship distribution}
\end{center}
\end{table}

Figure \ref{system_architecture}, represents the cybersecurity pipeline. Dark Caracal and CrossRat malware descriptions are provided as input to the pipeline. The resulting knowledge graph is shown in Figure \ref{populated kg} and more details about the resulting knowledge graph have been discussed in Section \ref{kgdc}.

We have 1287 pairs of entities from Dark Caracal malware description and 123 pairs from CrossRat malware description after filtering out the rest of entity pairs by methods mentioned in section \ref{sec RelExt}. Table \ref{relationships in docs} describes the relationships predicted by RelExt from the preprocessed input. Whereas, Table \ref{relationship distribution} shows various classes of relationships predicted for the Dark Caracal and CrossRat malware descriptions. 

\section{Conclusion}
\label{sec conclusion}
Cybersecurity Knowledge Graphs (CKG) are useful for storing a large number of semantic triples about cybersecurity entities. When we take an entity relationship set extracted by RelExt and assert it in a knowledge graph, we get access to significant information about various cybersecurity entities.

One such example is the use of Query Languages like SPARQL \cite{sparql} or SWRL \cite{swrl} to write queries which can be used to infer information about entities present in a knowledge graph. For example, we can use query languages, like SPARQL to ask the following question: \textit{What software have exposed vulnerabilities?}

\begin{lstlisting}
SELECT ?x WHERE {
?x :type :SOFTWARE.
?y :type :VULNERABILITY.
?x :hasVulnerability ?y.} 
\end{lstlisting}

The above query will return 'Adobe Flash Player' as an answer. 

In this paper, we develop the RelExt system that aids in the prediction of relationships between entity-pairs, using a neural network, specifically for cybersecurity. We describe a detailed pipeline, about how we prepare the corpus of cybersecurity texts, and process it further to provide as an input to our RelExt system. The preprocessing step removes entities which are not in proximity of each other as they occur in the cybersecurity text, or they cannot have a meaningful relationship between them. This removes a significant number of candidate semantic triples which are incorrect according to the STIX 2.0 schema. After training is complete, and our RelExt system has learned how to predict a relationship between pairs of entities, we use it to predict relationships from cybersecurity entities scraped from unseen cybersecurity text. This gives us a triple set for cybersecurity which can be asserted in a knowledge graph. 

The RelExt system can be used to aid in the development of an entity-relationship set specifically for cybersecurity, which can then be asserted in a cybersecurity knowledge graph. 
We achieved an accuracy of 96.61 \% over various data splits. RelExt successfully predicted over 700 relationships from Dark Caracal and CrossRat malware descriptions.

\bibliographystyle{plain}
\bibliography{malware}

\end{document}